\documentclass[twocolumn,usenames,dvipsnames]{autart}    

\usepackage{graphicx}          
\makeatletter \let\cl@part\relax \makeatother 
\usepackage{amsmath} 
\usepackage{amssymb}
\usepackage{balance}
\usepackage{mathtools}
\usepackage{siunitx}
\usepackage[scaled]{helvet}
\usepackage[usenames,dvipsnames]{xcolor}
\usepackage{pgfplots}
\usepackage{pgfplotstable}
\usepgfplotslibrary{fillbetween}
\usepackage{tabularx}
\usepackage{booktabs}
\usepackage{filecontents}
\pgfplotsset{compat = newest}
\usetikzlibrary{arrows,positioning,shapes,intersections,patterns,calc}
\usepgfplotslibrary{groupplots}
\usepackage{placeins}
\newtheorem{propy}{Property}

\newcommand\tran{\mkern-2mu\raise1.25ex\hbox{$\scriptscriptstyle\top\hspace{0.5mm}$}\mkern-3.5mu}
\newcommand{\R}{\mathbb{R}}
\newcommand{\N}{\mathbb{N}}

\newcommand{\bm}[1]{{\boldsymbol{#1}}}
\newcommand{\Verts}[1]{{\Vert #1 \Vert}}
\DeclareMathOperator{\diag}{diag}
\DeclareMathOperator{\var}{var}
\DeclareMathOperator{\mean}{\mu}
\DeclareMathOperator{\Var}{\Sigma}
\DeclareMathOperator{\Mean}{\bm\mu}

\newcommand{\g}{\bm g}
\newcommand{\ddqd}{\ddot{\bm q}_\bm{d}}
\newcommand{\dqd}{\dot{\bm q}_\bm{d}}
\newcommand{\qd}{\bm{q}_\bm{d}}
\newcommand{\ddqe}{\ddot{{\bm e}}}
\newcommand{\dqe}{\dot{{\bm e}}}
\newcommand{\qe}{{\bm e}}
\newcommand{\ddq}{\ddot{\bm q}}
\newcommand{\dq}{\dot{\bm q}}
\newcommand{\q}{\bm q}

\newcommand{\x}{\bm x}

\newcommand{\inputu}{\bm u}
\newcommand{\udyn}{\bm{f_u}}
\newcommand{\y}{\bm y}
\usepackage[noabbrev]{cleveref} 
\crefname{propy}{Property}{Properties}
\crefname{rem}{Remark}{Remarks}
\crefname{assum}{Assumption}{Assumptions}
\crefname{prop}{Proposition}{Propositions}
\crefname{cor}{Corollary}{Corollaries}
\crefname{lem}{Lemma}{Lemmas}
\crefname{thm}{Theorem}{Theorems}
\crefname{defn}{Definition}{Definitions}
\crefname{figure}{Fig.}{Fig.}
\Crefname{figure}{Figure}{Figures}
\crefname{equation}{}{}

\begin{document}
  \setlength{\abovedisplayskip}{5pt}
\setlength{\belowdisplayskip}{5pt}
\setlength{\abovedisplayshortskip}{5pt}
\setlength{\belowdisplayshortskip}{5pt}
\begin{frontmatter}

\title{Stable Gaussian Process based Tracking Control of Euler-Lagrange Systems} 

\thanks[footnoteinfo]{\copyright 2019. This manuscript version is made available under the CC-BY-NC-ND 4.0 license http://creativecommons.org/licenses/by-nc-nd/4.0/}

\author[a]{Thomas Beckers}\ead{t.beckers@tum.de},    
\author[b]{Dana Kuli\'c}\ead{dana.kulic@uwaterloo.ca}, 
\author[a]{Sandra Hirche}\ead{hirche@tum.de}               

\address[a]{Chair of Information-oriented Control (ITR), Department of Electrical and Computer Engineering,
Technical University of Munich, 80333 Munich, Germany}  

\address[b]{Adaptive Systems Laboratory, Department of Electrical and Computer Engineering, University of Waterloo, Waterloo, ON N2L 3G1, Canada}  
          

\begin{abstract}                         
	Perfect tracking control for real-world Euler-Lagrange systems is challenging due to uncertainties in the system model and external disturbances. The magnitude of the tracking error can be reduced either by increasing the feedback gains or improving the model of the system. The latter is clearly preferable as it allows to maintain good tracking performance at low feedback gains. However, accurate models are often difficult to obtain. \\
	In this article, we address the problem of stable high-performance tracking control for unknown Euler-Lagrange systems. In particular, we employ Gaussian Process regression to obtain a data-driven model that is used for the feed-forward compensation of unknown dynamics of the system. The model fidelity is used to adapt the feedback gains allowing low feedback gains in state space regions of high model confidence. The proposed control law guarantees a globally bounded tracking error with a specific probability. Simulation studies demonstrate the superiority over state of the art tracking control approaches.
\end{abstract}

\end{frontmatter}

\section{Introduction}
Euler-Lagrange (EL) systems represent a crucial and large class of dynamical systems, for which the equations of motion can be derived via the EL equation. Various control schemes for this class of systems have been proposed. Most of them can be considered as a subset of computed torque control laws (CTC), which enable very effective controllers in robust, adaptive and learning control~\cite{siciliano2010robotics}. The controller is separated into a feed-forward and a feedback part. A precise model of the true system is necessary to compensate the system dynamics to achieve a low gain feedback term. This is beneficial in many ways: it avoids large errors in the presence of noise~\cite{isidori2013nonlinear}, avoids the saturation of actuators~\cite{kapila2002actuator}, and enhances safety in applications such as human-robot interaction~\cite{de2008atlas}. Since the accuracy of the compensation depends on the precision of the model, all generalized external forces such as, e.g. in robotics, friction, payload or contact forces with the environment must be incorporated as precisely as possible. However, an accurate model of these uncertainties is hard to obtain by classical first principles based techniques. Especially in modern applications of Lagrangian systems such as service robotics, the interaction with unstructured and a priori unknown environments further increases the uncertainty. A common approach is to derive a dynamic model from first order physics and increase the feedback gains to compensate the uncertainties until a desired tracking performance is achieved~\cite{spong2006robot}. However, high gain control is undesirable (as explained above) and deriving a more accurate model of the system is often difficult if not impossible, e.g. in soft robotics~\cite{amiri2016control}. Additionally, the stability of the closed loop system might not be guaranteed. 

In this article, we address the problem of stable tracking control for EL systems with unknown dynamics. For this purpose, we use Gaussian Process regression (GPR), which is a data-driven learning approach. In particular, GPR is a supervised learning technique which combines several advantages. It requires only a minimum of prior knowledge to represent an arbitrary complex function, generalizes well even for small training data sets and has a precise trade-off between fitting the data and smoothing~\cite{rasmussen2006gaussian}. We employ the provided model confidence to adapt the feedback gains in areas where it is necessary to keep the system stable and the tracking error less than a given value. Computed torque control requires a parametric model of the EL system which can be identified, e.g. for robot manipulators~\cite{sciavicco2012modelling}. Errors in the identified dynamics deteriorate the tracking performance and can affect the stability of the closed loop. Several methods are presented to overcome this problem~\cite{slotine1987adaptive,ravichandran2004stability} but need an underlying parametric model. The idea to use GPR as a data-driven approach in control of robotic systems has been presented in~\cite{nguyen2008computed,alberto2014computed}. However, no stability guarantees are given. In~\cite{chowdhary2015bayesian,Berkenkamp2016ROA}, the stability of systems with GPR are analyzed but without exploiting the particular structure of EL systems and confidence-dependent feedback gains. Thus, high performance tracking control of EL systems with unknown dynamics and stability guarantees is still an open challenge.

\textbf{Contributions:} We develop a computed torque control law with GPR based feed-forward compensation (CTC-GPR) with an explicit tracking error computation. For this purpose, a GP learns the unknown system dynamics from training data. The proposed control law uses the mean of the GPR to compensate the unknown dynamics and the model confidence to adapt the gains. The derived method guarantees that the tracking error is ultimately bounded within a ball with a specific radius and a given probability. In the previous work~\cite{beckers:cdc2017} of the authors, first results for stable control of EL systems with GPR are presented. This article significantly extends the work by an explicit computation of the tracking error such that quantitative requirements for the closed loop performance can be imposed. Additionally, the work is enhanced by no need of a diagonal feedback gain matrix which allows to tune the performance over a wider range and fewer restrictions on the Lagrangian system, i.e. the generalized inertia matrix does not have to be bounded.

\section{Preliminaries and Definitions}\label{sec:def}
In this article, we focus on the class of non-conservative and fully-actuated systems where the equations of motion are given by\footnote{Vectors~$\bm a$ and vector-valued functions~$\bm f(\cdot)$ are denoted with bold characters. Matrices are described with capital letters. $I$ is the identity matrix and $0$ the zero matrix. The expression~$A_{:,i}$ denotes the i-th column of $A$. The smallest eigenvalue of a matrix is denoted by~$\underline{\lambda}(\cdot)$ and the largest by~$\bar\lambda(\cdot)$. The matrix norm is given by~$\Verts{A}=\bar\lambda(A\tran A)^{1/2}$.}
\begin{align}
	\frac{d}{dt}\left( \frac{\partial\mathcal{L} }{\partial \dq} \right)-\frac{\partial\mathcal{L} }{\partial \q}=\inputu_c+\inputu_{d}\label{for:el}
\end{align}
with the generalized coordinates~$\q\in\R^n$ and the general Lagrangian function~$\mathcal{L}(\dq,\q)\coloneqq\mathcal{T}(\dq,\q)-\mathcal{V}(\q)$. This function depends on the kinetic energy (or co-energy)~$\mathcal{T}\colon\R^n\times\R^n\to\R$ and the potential function~$\mathcal{V}\colon\R^n\to\R$. Two types of generalized external forces are considered: The action of control~$\inputu\in\R^{n}$ and the effect of the unknown dynamics~$\inputu_d\in\R^n$.
\begin{assum}\label{as:para}
	The unknown dynamics~$\inputu_d$ in~\cref{for:el} can be parametrized as $\inputu_d=\udyn\left(\bm{p}\right)$ with~$\bm{p}=[\ddq\tran,\dq\tran,\q\tran]\tran$ where~$\udyn\colon\R^{3n}\to\R^{n}$ is a continuous function.
\end{assum}
The assumption restricts~$\udyn$ to be not directly time dependent which holds in many application scenarios. For example, the common unknown dynamics in robotic systems, i.e. Columb and viscous friction, are included. The kinetic energy in the EL equation~\cref{for:el} is of the form~$\mathcal{T}(\dq,\q)=\frac{1}{2}\dq\tran H(\q)\dq$ where~$H(\q)\colon\R^n\to\R^{n\times n}$ is the symmetric and positive definite generalized inertia matrix.
Based on these assumptions, the EL equations~\cref{for:el} can be written in the equivalent form
\begin{align}
	\label{for:dyn_model_man}
	H(\q)\ddq+C(\q,\dq)\dq+\bm \g(\q)-\udyn(\bm{p})=\inputu(t),
\end{align}
where~$C(\q,\dq)\colon\R^n\times \R^n\!\to\!\R^{n\times n}$ is the generalized Coriolis matrix and the generalized vector~$\bm \g(\q)\colon\R^n\!\to\!\R^n$ is given by $\bm \g(\q)\!\coloneqq\frac{\partial\mathcal{V}(\q)}{\partial\q}$. The time-dependency of the states is omitted for simplicity of notation and the time-dependency of the input~$\inputu\colon\R_{\geq 0}\to\R^{n}$ might be also indirect, i.e. $\inputu(\bm{p}(t))$.
\begin{rem}\label{propy:structural}
	In this paper, the non-unique matrix~$C(\q,\dq)$ is always defined such that~$\dot{H}(\q)-2C(\q,\dq)\in\R^{n\times n}$ is skew-symmetric~$\forall \dq,\q\in\R^n$ following~\cite[Lemma 4.2]{murray1994mathematical}.
\end{rem}
\subsection{Gaussian Process Regression}
Assume a vector-valued and nonlinear function~$\bm f(\x)$ with~$\bm f\colon \R^n\to \R^n,n\in\N$. The measurement~$\y\in\R^n$ of the function is corrupted by Gaussian noise~$\bm\eta\in\R^n$, such that $\y=\bm f(\x)+\bm\eta$ where $\bm\eta\sim\mathcal{N}(\bm 0,\diag (\sigma_{1}^2,\ldots,\sigma_{n}^2))$ with the standard deviation~$\sigma_{1},\ldots,\sigma_{n}\in\R_{\geq 0}$. The training data~$\mathcal D=\{X,Y\}$ consists of~$m$ function evaluations at~$X=[\bm x^{\{1\}},\bm x^{\{2\}},\ldots,\bm x^{\{m\}}]\in\R^{n\times m}$ with the output values~$Y=[\y^{\{1\}},\y^{\{2\}},\ldots,\y^{\{m\}}]\tran\in\R^{m\times n}$. The prediction of each component of~$\bm f$ at a test input~$\x^*\in\R^n$ is derived from a Gaussian joint distribution. The conditional Gaussian distribution is defined by the mean and the variance
\begin{align}
	\mean(f_i\vert \x^*,\mathcal D)&=\bm k_{\Phi_i}(\x^*,X)\tran (K_{\Phi_i}+I \sigma^2_{i})^{-1}Y_{:,i}\\
	\var(f_i\vert \x^*,\mathcal D)&=k_{\Phi_i}(\x^*,\x^*)-\bm k_{\Phi_i}(\x^*,X)\tran \notag\\
	& \phantom{{}=}(K_{\Phi_i}+I \sigma^2_{i})^{-1} \bm k_{\Phi_i}(\x^*,X)
\end{align}
with the covariance function~$k_{\Phi_i}\colon\R^n\times\R^n\to\R$ as a measure of the correlation of two points~$(\bm x,\bm x^\prime)$. The matrix function~$K_{\Phi_i}\colon\R^{n\times m}\times \R^{n\times m}\to\R^{m\times m}$ is called the covariance matrix $K_{j,l}= k_{\Phi_i}(X_{:, l},X_{:, j})$ with~$j,l\in\lbrace 1,\ldots,m\rbrace$ where each element of the matrix represents the covariance between two elements of the training data~$X$. The vector-valued covariance function~$\bm k_{\Phi_i}\colon\R^n\times \R^{n\times m}\to\R^m$ calculates the covariance between the test input~$\x^*$ and the training data~$X$, i.e. $k_{\Phi_i,j} = k_{\Phi_i}(\x^*,X_{:, j})$ for all~$j\in\lbrace 1,\ldots,m\rbrace$ and~$i\in\lbrace 1,\ldots,n\rbrace$. These functions depend on a set of hyperparameters~$\Phi_i=\{\varphi^{\{1\}}_i,\ldots,\varphi^{\{n_h\}}_i\}$ whose number $n_h\in\N$ and domain of parameters depends on the covariance function used. A comparison of the characteristics of the different covariance functions can be found in~\cite{bishop2006pattern}.\\
The~$n$ components of~$\bm f\vert \x^*,\mathcal D$ are combined into a multi-variable Gaussian distribution with
\begin{align}
	\bm \mean(\bm f\vert \x^*,\mathcal D)&=[\mean(f_1\vert \x^*,\mathcal D),\ldots,\mean(f_n\vert \x^*,\mathcal D)]\tran\label{for:multigp}\\
	\Var(\bm f\vert \x^*,\mathcal D)&=\diag(\var(f_1\vert \x^*,\mathcal D),\ldots,\var(f_n\vert \x^*,\mathcal D)),\notag
\end{align} 
where the hyperparameters~$\Phi_i$ are optimized by means of the likelihood function $\arg\max_{\varphi^{\{j\}}} \log P(Y_{:,i}|X,\varphi^{\{j\}})$ for all $i\in\{1,\ldots,n\}$ and $j\in\{1,\ldots,n_h\}$. For this purpose, a gradient based algorithm is often used to find a (local) maximum of the likelihood function~\cite{rasmussen2006gaussian}. The computation of the variance with respect to a subset of elements of~$\x^*$ can be done by marginalization. Assume~$\x^*=[\x^{*}_1\tran,\x^{*}_2\tran]\tran$ with~$\x^*_1\in\R^{n_1},\x^*_2\in\R^{n_2}$ and~$\x^*\in\R^{n=n_1+n_2}$. The marginal variance of the prediction based on~$\x^*_1$ is
\begin{align}
	\var(f_i\vert \x^*_1,\mathcal D)=k_{\tilde{\Phi}_i}&(\x^*_1,\x^*_1)-\bm k_{\tilde{\Phi}_i}(\x^*_1,X_{1:n_1,:})\tran \notag\\
	&(K_{\tilde{\Phi}_i}(X_{1:n_1,:},X_{1:n_1,:})+I \sigma^2_{i})^{-1}\notag\\
	\bm k_{\tilde{\Phi}_i}&(\x^*_1,X_{1:n_1,:})\label{for:margvar}
\end{align}
with the necessary subset~$\tilde{\Phi}_i\subset\Phi_i$ for the covariance function on the input space~$\R^{n_1}$. The combined marginal variance is $\Var(\y^*\vert \x^*_1,\mathcal D)=\diag(\var_1(\cdot),\ldots,\var_{n_1}(\cdot))$.
\section{Gaussian Process Model}
\label{sec:modeling}
\begin{figure}[b]
	\begin{center}
	 	\begin{tikzpicture}[auto, node distance=3cm,>=latex']
\tikzstyle{block} = [draw, fill=white, rectangle, minimum height=2em, minimum width=1em]
\tikzstyle{sum} = [draw, fill=white, circle, node distance=1cm]
\tikzstyle{input} = [coordinate]
\tikzstyle{output} = [coordinate]
\tikzstyle{t_output} = []
\tikzstyle{pinstyle} = [pin edge={to-,thin,black}]

    \node [input, name=input] {};
    \node [block, right of=input,node distance=1.9cm] (controller) {$\scriptstyle \textsf{Controller}$};
    \node [block, right of=controller,node distance=1.9cm] (robot) {$\scriptstyle \textsf{System } \cref{for:dyn_model_man}$};
    \node [block, right of=robot,node distance=3.1cm] (hatrobot) {$\scriptstyle \hat H\ddq+\hat C\dq+\hat \g$};
    
    \node [output, right of=hatrobot,node distance=1cm] (output) {};

	\draw [->] (input) -- node[name=t1] {$\scriptstyle \qd,\dqd,\ddqd$} (controller);
	\draw [->] (controller) -- node[name=t,below] {$\scriptstyle \inputu$} (robot);
    \draw [->] (robot) -- node[name=q] {$\scriptstyle \ddq,\dq,\q$} (hatrobot);
    \draw [-] (hatrobot) -- node[name=o,below, right=0.5mm] {$\scriptstyle \hat \inputu$} (output);
    
    \coordinate [below of=q,node distance=0.9cm] (dp) {};
    \node [t_output, right of=dp,node distance=1cm] (x_output) {$\scriptstyle\{\bm{p}^{\{i\}}\}_{i=1}^{m}$};
    \node [t_output, above of=x_output,node distance=1.8cm] (y_output) {$\scriptstyle\{\tilde{\bm\tau}^{\{i\}}\}_{i=1}^{m}$};
    \draw [->] (q) |- (dp) -| (controller);

    \node [sum, above of=q,node distance=0.4cm,label={[label distance=0cm]20:-}] (sum) {};
    \node [output, above right of=sum,node distance=1cm] (output2) {\{Y\}};
    
    \draw [->] (dp) -- node[name=q1] {} (x_output);
    \draw [->] (sum) |- (y_output);
    
    \draw [->] (t) |- (sum);
    \draw [->] (output) |- (sum);

\end{tikzpicture}
	 	\vspace{-0.5cm}
		\caption{Diagram of the generation of the training data set $\mathcal D\!\!=\!\!\{\bm{p}^{\{i\}},\tilde{\bm\tau}^{\{i\}}\}_{i=1}^{m}$, where the output $\tilde{\bm\tau}$ is the difference between the real system, given by~\cref{for:dyn_model_man}, and an estimated parametric model.\label{fig:TD}}
	\end{center}
\end{figure}
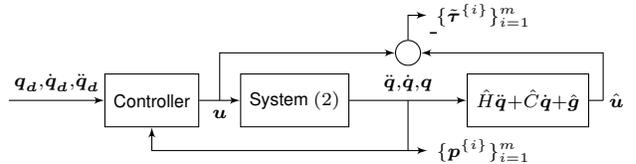

Consider the EL system in~\cref{for:dyn_model_man} with the unknown residual dynamics~$\udyn$. If a priori knowledge of the plant is available, a hybrid learning approach can be used which is a combination of a parametric and a data-driven model. We consider the estimated model to be given by
\begin{align}
	\hat \inputu(t)&=\hat H(\q)\ddq+\hat C(\q,\dq)\dq+\hat \g(\q),\label{for:learn}
\end{align}
where~$\hat H(\q)\in\R^{n \times n}, \hat C(\q,\dq)\in\R^{n \times n}$ and~$\hat\g(\q)\in\R^{n}$ are estimates of the true values which also satisfy~\cref{propy:structural}. Furthermore, the estimates must fulfill the following property.
\begin{propy}[Structure of the estimates]\label{propy:structural_est}
There exist~$h_1,h_2,k_C\!\in\!\R_{>0}$ with~$h_1\Verts{\x}^2\!\leq\!\x\tran \hat H(\q)\x\!\leq\! h_2\Verts{\x}^2$, and $\Vert \hat C(\q,\dq) \Vert\leq k_C\Vert \dq \Vert$ where~$\hat C(\q,\dq)\q^\prime=\hat C(\q,\q^\prime)\dq$ for all~$\q,\dq,\q^\prime,\x\in\R^n$.
\end{propy}
The identification of these estimates while guaranteeing~\cref{propy:structural,propy:structural_est} can be achieved following the identification procedures from~\cite{spong2006robot,kozlowski2012modelling}. Please note that~\cref{propy:structural_est} is required for the estimates only and not for the true system~\cref{for:dyn_model_man}, in comparison to~\cite{beckers:cdc2017}. 
\begin{rem}
	Without prior knowledge of the system, the estimates are set to~$\hat H=I,\hat C=0,\hat \g=\bm 0$.
\end{rem}
After the parametric model is selected, a GP is trained with~$m$ data pairs~$\mathcal D=\{\bm{p}^{\{i\}},\tilde{\bm\tau}^{\{i\}}\}_{i=1}^{m}$ of the system consisting of $\bm p=\left[\ddq\tran,\dq\tran,\q\tran\right]\tran\in\R^{3n}$ as input data, and the difference between the real system dynamics~\cref{for:dyn_model_man} and the estimated model~\cref{for:learn} as output data. This residual dynamic is denoted by
	\begin{align}
		\tilde{\bm \tau}(\bm p) &=\tilde H(\q)\ddq+\tilde C(\q,\dq)\dq+\tilde \g(\q)-\udyn(\bm{p}),\label{for:trainingset}
	\end{align}
with~$\tilde H=H-\hat H$,~$\tilde C=C-\hat C$ and~$\tilde{\g}=\g-\hat \g$. For the generation of training data, the system~\cref{for:dyn_model_man} can be operated by an arbitrary controller as shown in~\cref{fig:TD}. The only condition is that a finite sequence of training data of the system can be collected whereas stability is not necessarily required.
\vspace{-0.1cm}
\subsection{Model error}
\vspace{-0.1cm}
For the computation of the model error, we assume the following for the covariance function of the GP.
\vspace{-0.05cm}
\begin{assum}\label{as:rkhs}
	The covariance function~$k$ is chosen such that the functions~$\tilde{\tau}_1,\ldots,\tilde{\tau}_n$ have a bounded reproducing kernel Hilbert Space (RKHS) norm on any compact set~$\Omega\subset\R^{3n}$, i.e.~$\Verts{\tilde{\tau}_i}_k<\infty$ for all~$i=1,\ldots,n$.
\end{assum}
\vspace{-0.05cm}
\begin{rem}
	The norm of a function~$\bm f$ in a RKHS is a smoothness measure relative to a covariance function~$k$ that is uniquely connected with this RKHS. In particular, it is a Lipschitz constant with respect to the metric of the used covariance function. A more detailed discussion about RKHS norms is given in~\cite{wahba1990spline}. 
\end{rem}
\Cref{as:rkhs} requires that the covariance function must be selected in such a way that the residual~$\tilde{\bm \tau}(\bm{p})$ is an element of the associated RKHS. This sounds paradoxical since the residual is unknown. However, there exist some covariance functions, so called universal functions, which can approximate any continuous function arbitrarily precisely on a compact set~\cite[Lemma 4.55]{steinwart2008support}. Therefore, any smooth residual dynamics can be covered by a universal covariance function, i.e. this assumption is not restrictive. An upper bound for the distance between the mean prediction~$\Mean(\tilde{\bm \tau})$ of the GPR and the true function is given in \cite{srinivas2012information} and is extended for multidimensional functions in the following lemma.
\begin{lem}
	\label{lemma:boundederror}
	Consider a Lagrangian system~\cref{for:dyn_model_man} and a trained GP satisfying~\cref{as:rkhs}. The model error is bounded by
	\begin{align}
		\text{P}&\left\lbrace\Verts{\Mean(\tilde{\bm \tau}\vert \bm{p},\mathcal D)-\tilde{\bm \tau}(\bm p)}\leq \Verts{\bm \beta\tran \Var^{\frac{1}{2}}(\tilde{\bm \tau}\vert \bm{p},\mathcal D)}\right			\rbrace\geq \delta\label{for:upbound}
	\end{align}
	for~$\bm{p}\in \Omega$ with~$\delta\in(0,1),\bm\beta,\bm\gamma\in\R^n$ and
	\begin{align}
		\beta_j&=\sqrt{2\Verts{\tilde{\tau}_j}^2_k+300 \gamma_j \ln^3\left(\frac{m+1}{1-\delta^{1/n}}\right)}\notag\\
		\gamma_j&=\max_{\bm{p}^{\{1\}},\ldots,\bm{p}^{\{m+1\}}\in \Omega}\frac{1}{2}\log \vert I+\sigma_j^{-2}K_{\Phi_j}(\x,\x^\prime)\vert\notag\\
		\x,\x^\prime&\in\left\lbrace\bm{p}^{\{1\}},\ldots,\bm{p}^{\{m+1\}}\right\rbrace\notag
	\end{align}
\end{lem}
\vspace{-0.5cm}
\begin{pf}
	See~\cref{app:boundederror}.
\end{pf}
\begin{rem}
If~\cref{as:rkhs} is not fulfilled due to the wrong choice of covariance function or hyperparameters, for many common covariance functions the model error is still bounded on a compact set~\cite{beckers:cdc2016}. However, this may result in looser upper bounds for the model error. Tighter bounds might be achieved by using~\cite{beckers:cdc2018}.
\end{rem}
The information capacity~$\bm\gamma$ has a sub-linear dependency on the number of training points for many commonly used covariance functions~\cite{srinivas2012information}. Therefore, even though the values of the elements of~$\bm \beta$ are increasing with the number of training data, it is possible to learn the true function~$\tilde{\bm \tau}(\bm p)$ arbitrarily exactly~\cite{Berkenkamp2016ROA}. The result of~\cref{lemma:boundederror} is an upper bound for the model error. The stochastic nature of the bound is due to the fact that just a finite number of noisy training points are available. Since the model is used for a feed-forward compensation of the unknown dynamics of the system, the model error directly effects the tracking error as shown in the next section.
\section{Tracking control with GPR}\label{sec:ctrl_law}
The goal of tracking control is to follow a desired trajectory with the closed loop system. We start with the following assumption for the desired trajectory.
\begin{assum}
	\label{as:tra}
	The desired state trajectory is bounded by~$\Vert\qd\Vert\leq \bar{q}_d$,~$\Vert\dqd\Vert\leq \bar{\dot q}_d$ with~$\bar{q}_d,\bar{\dot q}_d\in\R_{\geq 0},\qd\in\R^n$.
\end{assum}
A bounded reference motion trajectories is a very natural assumption and does not pose any restriction in practice. Before the control law is proposed, the following assumption of the feedback gain functions $K_d$ and $K_p$ is introduced.
\begin{assum}
	\label{def:A}
	Let the functions~$\Var_d:\R^n\times\R^n\to\R^{n\times n}$ and~$\Var_p:\R^n\to\R^{n\times n}$ be the marginal variances which are defined analogously to~\cref{for:margvar} by $\Var_d(\dq,\q)\coloneqq\Var(\tilde{\bm \tau}\vert \dq,\q,\mathcal D)$ and $\Var_p(\q)\coloneqq\diag(\var(\tilde{\bm \tau}\vert q_1,\mathcal D),\ldots,\var(\tilde{\bm \tau}\vert q_n,\mathcal D))$.\\
	Let~$K_d,K_p\!\colon\!\R^{n\times n}\!\!\rightarrow\!\R^{n\times n}$ be symmetric matrix functions such that~$K_p(\Var_p(\q))\!=\!\diag(\Var_{p,11}(\q),\ldots,\Var_{p,nn}(\q))+K_c$ with~$K_c\in\R^{n\times n}$. The compositions~$(K_d\circ \Var_d)$,~$(K_p\circ \Var_p)$ are continuous and bounded by
	\begin{align}
		k_{d1}\Vert \x \Vert^2&\leq \x\tran K_d(\Var_d(\dq,\q))\bm x\leq k_{d2}\Vert \x \Vert^2\\
		k_{p1}\Vert \x \Vert^2&\leq \x\tran K_p(\Var_p(\q))\bm x\leq k_{p2}\Vert \x \Vert^2,
	\end{align}
	for all~$\dq,\q,\x\in\R^n$ with~$k_{p1},k_{p2},k_{d1},k_{d2}\in\R_{>0}$. 
\end{assum}	
\begin{rem}
\vspace{-0.1cm}
The feedback gains depend on the variance of the GP to adapt the gains based on the model confidence. We use the marginal variance such that the function $K_p$ implicitly depends exclusively on $\q$ and $K_d$ on $\q,\dq$ which is a common approach for variable feedback gains~\cite{ravichandran2004stability}. 
\end{rem}
\vspace{-0.2cm}
The next theorem introduces the control law with guaranteed boundedness of the tracking error.
\vspace{-0.1cm}
\begin{thm}[CTC-GPR]
	\label{theo:main}
	Consider the Lagrangian system~\cref{for:dyn_model_man} and a GP trained with~\cref{for:trainingset} which satisfies~\cref{as:para,as:rkhs}.
	Let~$\qe=\q-\qd$ be the tracking error with~\cref{as:tra} satisfied. The control law
	\begin{align}
		\inputu(t)&=\hat H(\q)\ddqd+\hat C(\q,\dq)\dqd+\hat \g(\q)+\Mean(\tilde{\bm \tau}\vert \bm{p},\mathcal D)\notag\\
		&-K_d(\Var_d) \dqe-K_p(\Var_p) \qe\label{for:control_law}
	\end{align}
	guarantees that there exist a compact set $\Omega$ and a model error $\bar\Delta$ such that
	\begin{align}
		P\left\lbrace\Vert\dqe(t),\qe(t)\Vert\leq r,\forall t\geq t_0+T(\delta)\right\rbrace\geq \delta\label{for:ball}
	\end{align}
	for any~$\Verts{\dqe\tran(t_0),\qe\tran(t_0)}<\delta_0$ with~$t_0,T(\delta_0),\delta_0,r\in\R_{>0}$.
\end{thm}
\vspace{-0.1cm}
Before proving the theorem we provide a series of results on a suitable Lyapunov candidate adapted from~\cite{ravichandran2004stability}.
\vspace{-0.1cm}
\begin{lem}
	\label{lemma:lyap}
	There exist an~$\varepsilon>0$ such that
	\begin{align}
		V\!\!=\!\!\frac{1}{2}\dqe\tran \hat H(\q)\dqe\!+\!\!\!\int_0^\qe\!\!\!\! \bm z\tran K_p(\Var_p(\bm z\!+\!\qd)\!)d\bm z\!+\!\varepsilon \qe\tran \hat H(\q)\dqe\label{for:lyap}
	\end{align}
	is a radially unbounded Lyapunov function.
\end{lem}
\vspace{-0.5cm}
\begin{pf}
	To ensure that the Lyapunov candidate is positive definite, the domain of the integral in~\cref{for:lyap} is analyzed. The integral is lower bounded by 
	\begin{align}
	\int_0^\qe\!\!\! \bm z\tran K_p(\Var_p)d\bm z\!&\geq\! \int_0^\qe \bm z\tran I\hspace{-2ex}\min_{i=\{1,\ldots,n\}}\hspace{-2ex}\lambda_i(K_p(\Var_p))d\bm z\\
	&\geq \!\frac{1}{2}\qe\tran\!\qe\hspace{-4ex}\min_{\hspace{1ex}\q\in\R,i=\{1,\ldots,n\}}\hspace{-5ex}\lambda_i(K_p(\Var_p(\q)))\!\geq\! \frac{1}{2} k_{p1} \Vert \qe \Vert^2,\notag
	\end{align}	
	where $\lambda_i$ denote the eigenvalues of the matrix $K_p(\cdot)$. An upper quadratic bound can be found in an analogous way using the maximum eigenvalue of $K_p(\cdot)$. Since the integral is lower bounded and~$\hat H(\q)$ is always positive definite, the parameter~$\varepsilon$ can be chosen sufficiently small to achieve a positive definite and radially unbounded Lyapunov function. The valid interval for~$\varepsilon$ can be determined by the lower bound of the Lyapunov function~\cref{for:lyap}
	\begin{align}
		V(\dqe,\qe)\geq \frac{1}{2}h_1\Verts{\dqe}^2+\frac{1}{2}k_{p1}\Verts{\qe}^2-\frac{1}{2}\varepsilon h_2\left( \Verts{\dqe}^2+\Verts{\qe}^2\right)\notag
	\end{align}
	which is positive for $0<\varepsilon < \min\left\lbrace k_{p1}/h_2,h_1/h_2\right\rbrace$.\qed
\end{pf}
\vspace{-0.5cm}
In the next step, we derive an upper bound for the time derivative of the Lyapunov function.
\begin{lem}
	\label{lemma:boundLyap}
	Consider the Lyapunov function~\cref{for:lyap} and the system~\cref{for:dyn_model_man} with the control law~\cref{for:control_law}. The drift of~\cref{for:lyap} is upper bounded with probability $ \delta\in(0,1)$ by
	\begin{align}
	\text{P}\big\lbrace\dot V&\leq -\frac{3}{4}v_1 \Verts{\dqe}^2-\frac{3}{4}\varepsilon v_2 \Verts{\qe}+\varepsilon k_C\Verts{\dqe}^2\Verts{\qe}\notag\\
	&+ \frac{{\bar \Delta}^2 }{v_1}+ \varepsilon\frac{{\bar \Delta}^2 }{v_2}\big\rbrace\geq \delta\label{for:boundLyap}\\
		\begin{split}
	v_1&:=-\varepsilon h_2+k_{d1}-\frac{\varepsilon \rho}{2}(k_C \bar{\dot q}_d+k_{d2})\\
	v_2&:= k_{p1}\frac{\varepsilon_2}{1+\varepsilon_2}.
	\label{for:def_v}
	\end{split}\\
	0<\varepsilon &<\min\left\lbrace \frac{k_{p1}}{h_2},\frac{h_1}{h_2},\frac{2 k_{d1}}{2 h_2+\rho(k_C \bar{\dot q}_d+k_{d2})}\right\rbrace\label{for:epsilon2}
	\end{align}
	with~$v_1,v_2,\bar \Delta\in\R_{>0}$ and $\bar \Delta\geq\Verts{\bm \beta\tran \Var^{\frac{1}{2}}(\tilde{\bm \tau}\vert \bm{p},\mathcal D)},\forall\bm p\in \Omega$.
\end{lem}
\vspace{-0.5cm}
\begin{pf}
The time derivative of~\cref{for:lyap} is expressed by
	\begin{align}
		\dot V\!=\!\begin{bmatrix}\dqe\tran \hat H+\varepsilon\qe\tran \hat H\\ \qe\tran K_p(\Var_p) +\frac{1}{2} \dqe\tran \dot{\hat H}+\varepsilon (\qe\tran \dot{\hat H}+ \dqe\tran \hat H)\end{bmatrix}^\top\!\!\begin{bmatrix} \ddqe\\ \dqe \end{bmatrix}\label{for:lyapdev},
	\end{align}
	using the symmetry of $\hat H$ and 
	\begin{align}	
		\frac{\partial }{\partial \qe}\int_{\bm 0}^\qe \bm z\tran K_p(\Var_p(\bm z+\qd))d\bm z=\qe\tran K_p(\Var_p(\q)).
	\end{align}	
	For the computation of $\ddqe$, the closed loop dynamics for the EL system~\cref{for:dyn_model_man} with input~\cref{for:control_law} is determined by
	\begin{align}
		\ddqd={\hat H}^{-1}\left(H\ddq+C\dq+\bm \g-\udyn(\bm{p})-\hat C\dqd\notag\right.\\
		\left.-\hat \g-\Mean(\tilde{\bm \tau}\vert \bm{p},\mathcal D)+K_d(\Var_d) \dqe+K_p(\Var_p) \qe  \right)\label{for:qdforV}.
	\end{align}
	With $\hat{C}\dqd=\hat{C}\dq-\hat{C}\dqe$ and \cref{for:trainingset}, the closed loop dynamics is rewritten as
	\begin{align}
		\ddqe=\ddq-\ddqd&={\hat H}^{-1}\big(\Mean(\tilde{\bm \tau}\vert \bm{p},\mathcal D)-\tilde{\bm \tau}(\bm p)\notag\\
		&-K_d(\Var_d) \dqe-K_p(\Var_p)\qe-\hat C\dqe  \big).
	\end{align}	
	Using the last equation and~\cref{for:lyapdev}, the time derivative of the Lyapunov function~\cref{for:lyap} is expressed by
	\begin{align}
		\dot V=&\begin{bmatrix} \dqe\\\qe \end{bmatrix}^\top\underbrace{\begin{bmatrix}
		\underbrace{\vphantom{\frac{\varepsilon}{2}}-K_d(\Var_d)+\varepsilon \hat H}_{M_{11}} & \underbrace{\frac{\varepsilon}{2}(- K_d\tran(\Var_d)+ \hat C)}_{M_{12}}			\\
		\underbrace{\frac{\varepsilon}{2}(- K_d(\Var_d)+ \hat C\tran)}_{M_{12}} & \underbrace{\vphantom{\frac{\varepsilon}{2}}- \varepsilon K_p(\Var_p)}_{M_{22}}
		\end{bmatrix}}_{M\in\R^{2n\times 2n}}\begin{bmatrix} \dqe\\\qe \end{bmatrix}\notag\\
		+&(\dqe+\varepsilon\qe)\tran(\Mean(\tilde{\bm \tau}\vert \bm{p},\mathcal D)-\tilde{\bm \tau}(\bm p)),\label{for:dotV}
	\end{align}
	where the skew-symmetry of $\dot{\hat{H}}-2\hat{C}$ is exploited.
	For the following analysis, we compute bounds for the elements of the matrix~$M$ to bound the drift of the Lyapunov function, based on~\cite{ravichandran2004stability}. The matrix~$M_{11}\in\R^{n\times n}$ is negative definite for sufficiently small~$\varepsilon>0$ and upper bounded with $\dqe\tran M_{11}\dqe\leq(-k_{d1}+\varepsilon h_2)\Verts{\dqe}^2$. Analogously, the submatrix~$M_{22}\in\R^{n\times n}$ is negative definite with~$\qe\tran M_{22}\qe\leq-\varepsilon k_{p1}\Vert \qe\Vert^2$. With~\cref{as:tra} and~\cref{propy:structural_est}, the submatrix~$M_{12}\in\R^{n\times n}$ is upper bounded by
	\begin{align}
		\qe\tran M_{12} \dqe\leq\varepsilon \left( k_C\Vert\dqe\Vert+k_C \bar{\dot q}_d+ k_{d2}\right) \Vert \dqe \Vert \Vert \qe \Vert
	\end{align}
	With~\cref{lemma:boundederror}, the overall upper bound for the time derivative of the Lyapunov function is given by
	\begin{align}
		\text{P}\big\lbrace\dot V&\leq (\varepsilon h_2-k_{d1}) \Vert \dqe \Vert^2-\varepsilon k_{p1}\Vert \qe\Vert^2\notag\\
		&+\varepsilon \left( k_C\Vert\dqe\Vert+k_C \bar{\dot q}_d+ k_{d2}\right) \Verts{\dqe}\Verts{\qe}\label{for:Vbound}\\
		&+(\Verts{\dqe}+\varepsilon\Verts{\qe})\Verts{\bm \beta\tran \Var^{\frac{1}{2}}(\tilde{\bm \tau}\vert \bm{p},\mathcal D)} \big\rbrace\geq \delta.\notag
	\end{align}
	For the next step, we consider the Peter-Paul inequality given by $\Verts{\dqe}\Verts{\qe}\leq \frac{1}{2}\left( \rho \Verts{\dqe}^2+\Verts{\qe}^2/\rho\right)$ that holds for all~$\dqe,\qe\in\R^n$ and~$\rho\in\R_{\geq 0}$,~\cref{for:Vbound} can be rewritten as
	\begin{align}
		\text{P}\big\lbrace&\dot V\leq (\varepsilon h_2-k_{d1}) \Vert \dqe \Vert^2-\varepsilon k_{p1}\Vert \qe\Vert^2\notag\\
		&+\frac{\varepsilon}{2} \left(k_C \bar{\dot q}_d+ k_{d2}\right)\!\!\left(\!\rho\Verts{\dqe}^2+\frac{\Verts{\qe}^2}{\rho}\!\right)\!\!+\varepsilon k_C\Verts{\dqe}^2\Verts{\qe}\notag\\
		&+(\Verts{\dqe}+\varepsilon\Verts{\qe})\Verts{\bm \beta\tran \Var^{\frac{1}{2}}(\tilde{\bm \tau}\vert \bm{p},\mathcal D))}\big\rbrace\!\geq\! \delta\label{for:Vbound2}\\
		&\text{with }\rho=(1+\varepsilon_2)\frac{k_C \bar{\dot q}_d+ k_{d2}}{2 k_{p1}},\,\varepsilon_2\in\R_{> 0}.\notag
	\end{align}
	The choice of~$\rho$ guarantees that the factors of the quadratic parts are still negative:
	\begin{align}
		\text{P}\big\lbrace\dot V&\leq \left(\varepsilon h_2-k_{d1}+\frac{\varepsilon \rho}{2}(k_C \bar{\dot q}_d+k_{d2})\right) \Verts{\dqe}^2\notag\\
		&-\varepsilon k_{p1}\frac{\varepsilon_2}{1+\varepsilon_2}\Verts{\qe}^2+\varepsilon k_C\Verts{\dqe}^2\Verts{\qe}\label{for:Vbound3}\\
		&+(\Verts{\dqe}+\varepsilon\Verts{\qe})\Verts{\bm \beta\tran \Var^{\frac{1}{2}}(\tilde{\bm \tau}\vert \bm{p},\mathcal D)}\big\rbrace\geq \delta\notag
	\end{align}
	Since the covariance function is continous and thus bounded on a compact set~$\Omega$, the variance~$\Var(\tilde{\bm \tau}\vert \bm{p},\mathcal D)$ is bounded, for more details see~\cite{beckers:cdc2016}. Thus, there exists an upper bound~$\bar \Delta$ for the model error. Applying the inequality $v_1\Verts{\x}\leq v_1^2/v_2+v_2\Verts{\x}^2/4$ that holds~$\forall\x\in\R^n$ and~$v_1,v_2\in\R_{\geq 0}$, the model error in~\cref{for:Vbound3} can be bounded by a quadratic function which results in~\cref{for:boundLyap}. The restrictions for $\varepsilon$ must be extended to~\cref{for:epsilon2} to ensure that the variables~$v_1,v_2\in\R_{>0}$ are positive.\qed
\end{pf}
\vspace{-0.6cm}
We are now ready to provide the proof of~\cref{theo:main}.
\vspace{-0.5cm}
\begin{pf*}{PROOF (\Cref{theo:main}).}
	According to~\cite[Theorem 1]{ravichandran2004stability} and~\cref{lemma:boundLyap,lemma:lyap}, there exists a~$\xi\in\R_{\geq 0}$ and~$\varrho\in\R_{\geq 0}$ for~\cref{for:Vbound} such that
	\begin{align}
		\text{P}\left\lbrace\dot V(\x,t)\leq -\xi V(\x,t)+\varrho\right\rbrace\geq \delta.
	\end{align}
	Consequently, using~\cite[Theorem 2.1]{corless1990guaranteed}, the closed loop is uniformly ultimately bounded and exponentially convergent to a ball with a probability of at least~$\delta$. Since the state is bounded, it is always possible to find a set $\Omega$ and a maximum model error $\bar\Delta$ such that $\bm{p}\in \Omega$. \qed
\end{pf*}
\vspace{-0.6cm}
Additionally, we can compute exactly the tracking error of the closed loop.
\vspace{-0.1cm}
\begin{prop}
The radius $r$ of the ball~\cref{for:ball} is 
	\begin{align}
		r&=\sqrt{\frac{2\varrho}{\xi\min\left\lbrace k_{p1}-\varepsilon h_2, h_1-\varepsilon h_2 \right\rbrace}}\label{for:radius}\\
		\xi&=\frac{2}{3}\frac{\min\left\lbrace\varepsilon v_2,v_1-\frac{4}{3}\varepsilon k_c \sqrt{\frac{ 2V_0}{k_{p1}-\varepsilon h_2}}\right\rbrace}{\max\left\lbrace\varepsilon h_2+k_{p2},(1+\varepsilon)h_2\right\rbrace}\label{for:xi}
	\end{align}
	with $\varrho={\bar \Delta}^2/v_1+ \varepsilon{\bar \Delta}^2/v_2$ where $v_1,v_2$ are defined by~\cref{for:def_v}, and $V_0=V(\bm 0,\bm 0)$. The restriction~\cref{for:epsilon2} is extended to
	\begin{align}
		0<\varepsilon < \min&\left\lbrace \frac{k_{p1}}{h_2},\frac{h_1}{h_2},\frac{2 k_{d1}}{2 h_2+\frac{2k_{p1}\rho^2}{1+\varepsilon_2}+\frac{8}{3}k_c\sqrt{\frac{ 2V_0}{k_{p1}-\varepsilon h_2}}}\right\rbrace .\notag
	\end{align}
\end{prop}
\vspace{-0.6cm}
\begin{pf}	
The proof follows from~\cite[Lemma 2.1]{wen1988new}.
\end{pf}	
\begin{rem}
If a perfect model was available, such that~$\bar\Delta=0$, equation~\cref{for:dotV} would show that the closed loop system is asymptotically stable because of the negative definiteness of $M$, see~\cref{app:mneg}.
\end{rem}
\vspace{-0.2cm}
\subsection{Design guidelines}
\vspace{-0.1cm}
\Cref{theo:main} provides an ultimate bound with a given probability depending on the gains, the system parameters and the variance of the GP. The radius of the bound depends quadratically on the upper bound of the model error~$\bar \Delta$. Thus, the radius~$r$ shrinks if the upper bound of the variance of the GPR decreases. The consequence is an improved tracking performance in terms of tracking error and the possibility to decrease the feedback gains which is beneficial for noise attenuation. The posterior variance of the GPR is related to the number and distribution of the training points and can be decreased, e.g., with the Bayesian optimization approach where the next training point is set to the position of maximum variance. For the commonly used squared exponential covariance, each new training point reduces the posterior variance~\cite{umlauft:cdc2017}.\\
The bounds of the adaptive gains also affect the radius of the ball. Increasing the lower bound of~$K_d$ shrinks the radius since~$\varepsilon$ can be arbitrarily small and $v_1$ depends linearly on $k_{d1}$. The influence of~$K_p$ depends on the Lagrangian system. Based on the results, different design goals can be addressed which are visualized in~\cref{fig:flow}.
\vspace{-0.2cm}
\section{Numerical Illustration}\label{sec:sim}
\vspace{-0.1cm}
In this section, we present examples\footnote{Source code: https://github.com/TBeckers/CTC\_GPR} illustrating the properties of the proposed CTC-GPR control scheme and a more detailed case study.
\subsection{Noise attenuation and saturation}
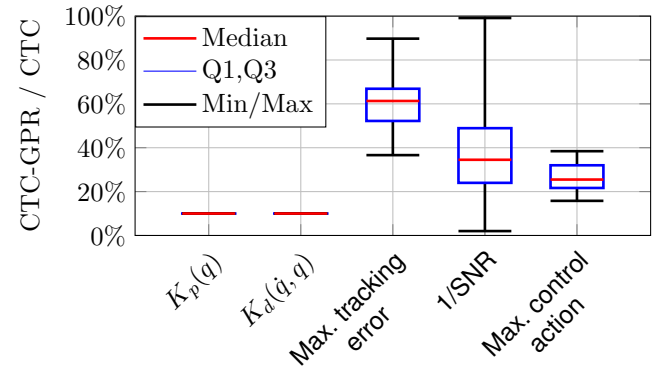
\begin{figure}[b]
	\begin{center}
	 	\begin{tikzpicture}
\pgfplotsset{
	every x tick label/.append style={align=center,rotate=45,anchor=north east},
    box plot/.style={
        /pgfplots/.cd,
        black,
        only marks,
        line width=1pt,
        mark options={line width=1pt},
        mark=-,
        mark size=1em,
        /pgfplots/error bars/.cd,
        y dir=plus,
        y explicit,
    },
    box plot box/.style={
        /pgfplots/error bars/draw error bar/.code 2 args={%
            \draw[line width=1pt]  ##1 -- ++(1em,0pt) |- ##2 -- ++(-1em,0pt) |- ##1 -- cycle;
        },
        /pgfplots/table/.cd,
        y index=2,
        y error expr={\thisrowno{3}-\thisrowno{2}},
        /pgfplots/.cd,
        blue,
        only marks,
        mark=-,
        mark size=1em,
        /pgfplots/error bars/.cd,
        y dir=plus,
        y explicit,
    },
    box plot top whisker/.style={
        /pgfplots/error bars/draw error bar/.code 2 args={%
            \pgfkeysgetvalue{/pgfplots/error bars/error mark}%
            {\pgfplotserrorbarsmark}%
            \pgfkeysgetvalue{/pgfplots/error bars/error mark options}%
            {\pgfplotserrorbarsmarkopts}%
            \path[line width=1pt] ##1 -- ##2;
        },
        /pgfplots/table/.cd,
        y index=4,
        y error expr={\thisrowno{2}-\thisrowno{4}},
        /pgfplots/box plot
    },
    box plot bottom whisker/.style={
        /pgfplots/error bars/draw error bar/.code 2 args={%
            \pgfkeysgetvalue{/pgfplots/error bars/error mark}%
            {\pgfplotserrorbarsmark}%
            \pgfkeysgetvalue{/pgfplots/error bars/error mark options}%
            {\pgfplotserrorbarsmarkopts}%
            \path[line width=1pt] ##1 -- ##2;
        },
        /pgfplots/table/.cd,
        y index=5,
        y error expr={\thisrowno{3}-\thisrowno{5}},
        /pgfplots/box plot
    },
    box plot median/.style={
        /pgfplots/.cd,
        red,
        line width=1pt,
        only marks,
        mark=-,
        mark size=1em,
        /pgfplots/error bars/.cd,
        y dir=plus,
        y explicit,
    }
}

\begin{axis} [grid, enlarge x limits=0.2,xtick=data,ytick={0,20,40,60,80,100},
			  ymin=0,ymax=100,
  			  height=4.5cm, width=\columnwidth,
			  xticklabels={$K_p(q)$,{$K_d(\dot{q},q)$},\textsf{Max. tracking}\\ \textsf{error},\textsf{1/SNR},\textsf{Max. control}\\ \textsf{action}},
			  ylabel={CTC-GPR / CTC},
			  yticklabels={0\%,20\%,40\%,60\%,80\%,100\%},
			  legend style={at={(0,1)},anchor=north west},
  		      legend cell align={left},reverse legend]
    \addplot [box plot top whisker] table {data/comp/comp_boxplot.dat};
    \addlegendentry{Min/Max}
    \addplot [forget plot,box plot bottom whisker] table {data/comp/comp_boxplot.dat};
        \addplot [box plot box] table {data/comp/comp_boxplot.dat};
    \addlegendentry{Q1,Q3}
    \addplot [box plot median] table {data/comp/comp_boxplot.dat};
    \addlegendentry{Median}
\end{axis}
\end{tikzpicture}
	 	\vspace{-1cm}
		\caption{Comparison between CTC and the proposed CTC-GPR for 30 randomly selected systems. CTC-GPR values are given as a percentage of CTC values.\label{fig:comp}}
	\end{center}
\end{figure}

In the following example, we show the benefit of the CTC-GPR in comparison to classical CTC. We assume a one dimensional EL system $\tau=\ddot{q}+\dot{q}+q+f_u(\bm{p})$ with 30 randomly generated dynamics
\begin{align}
f_u(\bm{p})=\frac{\dot{q}^2\sin(q-c)-\sin(c)}{\cos(q-c)-1.1\cos^{-1}(q-c)}
\end{align}
where each~$c$ is uniformly chosen from the set~$[0,2\pi]$. For the parametric model, we use the estimates~$\hat H=\hat C=\hat g=1$. The 441 training data pairs~$\{\tilde\tau\}$ and~$\{\ddq,\dq,\q\}$ for a GPR with squared exponential covariance function are equally distributed on the set~$[0]\times[-1,1]\times[-1,1]$. A conjugate gradient algorithm is used to minimize the log likelihood function to find suitable hyperparameters. The desired trajectory is given by~$\q_d=\sin(t)$ and the initial system value is~$q_0=0,\dot{q}_0=1$. The measurements of~$\ddot{q},\dot{q},q$ are corrupted by Gaussian noise with~$\mathcal{N}(0,0.04^2)$ for training and control. The simulation time is between zero and~$2\pi$ seconds. In the simulation, the CTC-GPR and the classical computed torque are compared in terms of the maximum tracking error, the noise attenuation and the maximum control action. The feedback gains of the CTC are~$K_p=100,K_d=100$ whereas the CTC-GPR is parameterized with
\begin{align}
K_p(q)&=10+100\Var_p(q)\\
K_d(\dot{q},q)&=10+100\Var_d(\dot{q},q).
\end{align}
The results are shown in~\cref{fig:comp}. The variation of the gains is minimal since the desired trajectory is inside the training area where the variance is quite low. The maximal tracking error~$\max\Verts{\dot{e}(t),e(t)}$ is decreased compared to CTC approach for all systems with a median of~$61.6\%$. The CTC-GPR shows remarkably better noise attenuation, as indicated by a higher signal to noise ratio (SNR) of the system trajectory. The SNR is computed as the ratio of the summed squared magnitude of the state to that of the noise. Also, the maximal control action is reduced due to the lower feedback gains of the CTC-GPR, which can prevent actuator saturation.
\subsection{Case study}
\vspace{-0.2cm}
In this case study, the benefit of the CTC-GPR is shown for a 2-link robotic manipulator~\cite[Page 164]{murray1994mathematical}. As reference for a performance comparison, we use CTC since most of the robotic control schemes can be considered as special cases of computed-torque controllers. We assume point masses for the links of~$m_1=m_2=\SI{1}{\kilogram}$, which are located in the center of each link. The length of the links is set to~$l_1=l_2=\SI{1}{\meter}$. The joints are without mass and not influenced by any friction. Gravity is assumed to be~$g=\SI{9.81}{\meter\second^{-2}}$. As estimates, we use~$\hat{m}_1=\SI{0.9}{\kilogram},\hat{m}_2=\SI{1.1}{\kilogram},\hat{l}_1=\SI{0.9}{\meter}$, and~$\hat{l}_2=\SI{1.1}{\meter}$. 
The initial joint angles are set to~$\q_0=[0,1]\tran$ and ~$\dq_0=[1,0]\tran$. The unknown dynamics~$\udyn(\bm{p})$ is simulated by an arbitrarily chosen function
\begin{align}
\renewcommand{\arraystretch}{0.95}
\udyn(\bm{p})=\begin{bmatrix}
\sin(2\dot{q}_2)+\cos(2q_1)+\ddot{q}_1\\
\sin(2\dot{q}_2)+2\sin(\dot{q}_1)
\end{bmatrix}.\label{for:exdyn}
\renewcommand{\arraystretch}{1}
\end{align}
A GP with a squared exponential covariance function learns the difference between the estimated model and the true system based on 576 equally distributed training pairs on the domain~$\q,\ddq\in[0,1]^2,\,\dq\in[-1,1]^2$. The measurements of~$\ddq,\dq,\q$ are corrupted by Gaussian noise with~$\mathcal{N}(0,0.1^2)$. The hyperparameters are optimized by means of the likelihood function. The desired trajectory is a sinusoidal function with~$\q_0=[0,1]\tran$. In this example, the gains are adapted with $K_p(\Var_p)=7I+400\Var_{p}(\q)$ and $
K_d(\Var_d)=6I+400\Var_{d}(\dq,\q)$. \Cref{fig:var_gains} shows the resulting trajectory for the first joint along with the desired trajectory. As comparison, we use a classic CTC with~$K_{p,s}=K_{d,s}=\diag(10,10)$ which is a trade-off between tracking error and high feedback gains. The advantages of the CTC-GPR with variable feedback gains in comparison to CTC are presented in~\Cref{tab:comp}. Additionally, this approach is compared to a CTC-GPR with static feedback gains where the values of the static gains are set to the minimum of the variable gains such that the noise attenuation is comparable.
\begin{figure}[t]
	\begin{center}
	 	\tikzstyle{decision} = [diamond, draw, fill=white,text width=4.5em, text badly centered, node distance=3cm, inner sep=0pt]
\tikzstyle{block} = [rectangle, draw, fill=white,text width=9em, text centered, rounded corners, minimum height=0.9cm]
\tikzstyle{wide_block} = [rectangle, draw, fill=white, text width=12em, text centered, rounded corners, minimum height=0.6cm,node distance = 1cm]
\tikzstyle{line_me} = [draw, -latex', draw=red,thick,dashed]
\tikzstyle{line_fg} = [draw, -latex', draw=blue,thick,dotted]
\tikzstyle{line_ra} = [draw, -latex',draw=green!70!black,thick]
\tikzstyle{cloud} = [draw, ellipse,fill=white, node distance=2cm, minimum height=2em]
    
\begin{tikzpicture}[node distance = 2cm, auto]
    \node [wide_block] (init) {Define estimated model};
    \node [wide_block, below of=init] (collect) {Collect training data};
    \coordinate [left of=collect,xshift=-2.2cm] (helper);
    \node [wide_block, below of=collect] (model error) {Compute model error};
    \node [block, below of=model error,node distance = 2.25cm and 2cm] (setf) {Set feedback functions};
    \node [block, below of=setf,node distance = 2.4cm and 2cm] (compr) {Compute radius};
    \node [block, below left=0.3cm and -1.7cm of model error] (setrf) {Set radius \& feedback functions};
    \node [block, below of=setrf,node distance = 2.4cm and 2cm] (compmax) {Compute max. allowed model error};
    \node [decision, below of=compmax,node distance = 2.6cm and 2cm] (decide) {Model accurate enough?};
    \node [block, below right=0.3cm and -1.7cm of model error] (setr) {Set radius};
    \node [block, below of=setr,node distance = 2.4cm and 2cm] (compb) {Compute bounds for feedback gains};
    \node [block, below of=compb,node distance = 2.4cm and 2cm] (des) {Design feedback functions};
    \node [cloud, below of=compr,node distance=2.75cm] (end) {End};
    \node [text width=8cm, below of=end,node distance = 0.6cm and 2cm] (text1) {Radius for specified feedback gain functions};
    \coordinate [left of=text1,xshift=-2.5cm] (htext1);
    \node [text width=8cm, below of=text1,node distance = 0.5cm and 2cm] (text2) {Sufficiently accurate model for predefined radius};
    \coordinate [left of=text2,xshift=-2.5cm] (htext2);
    \node [text width=8cm, below of=text2,node distance = 0.5cm and 2cm] (text3) {Feedback gain functions for predefined radius};
    \coordinate [left of=text3,xshift=-2.5cm] (htext3);
    \path [line_ra] (init) -- (collect);
    \path [line_me, transform canvas={shift={(-0.5cm,0)}}] (init) -- (collect);
    \path [line_fg, transform canvas={shift={(0.5cm,0)}}] (init) -- (collect);
    \path [line_ra] (collect) -- (model error);
    \path [line_me, transform canvas={shift={(-0.5cm,0)}}] (collect) -- (model error);
    \path [line_fg, transform canvas={shift={(0.5cm,0)}}] (collect) -- (model error);
    \path [line_ra] (model error) -- (setf);
    \path [line_ra] (setf) -- (compr);
    \path [line_ra] (compr) -- (end);
    \path [line_ra] (htext2) -- (text2);
    \path [line_me] (model error) -- (setrf);
    \path [line_me] (setrf) -- (compmax);
    \path [line_me] (compmax) -- (decide);
    \path [line_me,-] (decide) -| node [near start] {no} (helper);
    \path [line_me] (helper) -- node {more data} (collect);
    \path [line_me] (decide) |- node[xshift=0.8cm] {yes} (end);
    \path [line_me] (htext1) -- (text1);
    \path [line_fg] (model error) -- (setr);
    \path [line_fg] (setr) -- (compb);
    \path [line_fg] (compb) -- (des);
    \path [line_fg] (des) |- (end);
    \path [line_fg] (htext3) -- (text3);
\end{tikzpicture}
	 	\vspace{-0.65cm}
		\caption{Guidelines for different design goals.\label{fig:flow}}
		\vspace{-0.1cm}
	\end{center}
\end{figure}
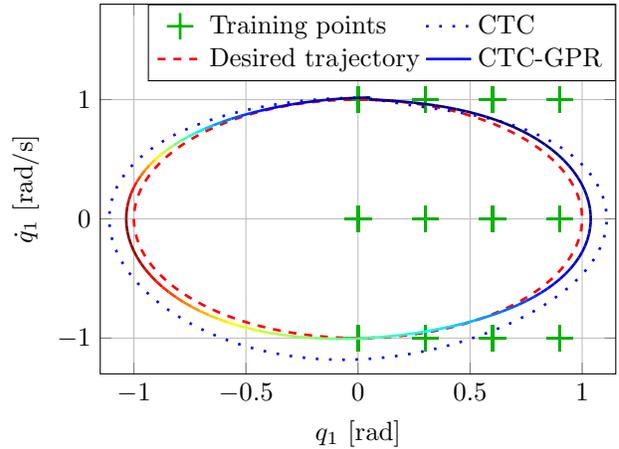
\begin{figure}[t]
\begin{center}
 \vspace{0.0cm}
 \begin{tikzpicture}
\pgfplotsset{
  set layers,
  mark layer=axis tick labels
}
\pgfplotsset{select coords between index/.style 2 args={
    x filter/.code={
        \ifnum\coordindex<#1\def\pgfmathresult{}\fi
        \ifnum\coordindex>#2\def\pgfmathresult{}\fi
    }
}}
\begin{axis}[
  xlabel={$q_1$ [rad]},
  ylabel={$\dot{q}_1$ [rad/s]},
  line width=1pt,
  grid = major,
  colormap/jet,
    height=6.5cm,
  width=\columnwidth,
  xmin=-1.15, xmax=1.15, ymin=-1.3, ymax=1.8,
  legend columns=2,legend style={at={(1,1)},anchor=north east},
  legend cell align={left},transpose legend]
\addplot[mark=+,color=green!70!black, only marks,mark size=5,line width=1pt] table [x index=0,y index=1]{data/casestudy/training_points.dat};
\addplot+[color=red,style=dashed,no marks,line width=1pt,select coords between index={1}{130}] table [x index=0,y index=1]{data/casestudy/classic_gains.dat};
\addplot+[color=blue,style=loosely dotted,no marks,line width=1pt,select coords between index={1}{130}] table [x index=2,y index=3]{data/casestudy/classic_gains.dat};
\addplot+[forget plot,mesh,point meta=\thisrowno{4}, no marks,line width=1pt, shader=interp,select coords between index={1}{128}] table [x index=2,y index=3]{data/casestudy/var_gains.dat};
\addplot+[color=blue,no marks,line width=1pt] coordinates {(3,3) (2,2)};
\legend{Training points,Desired trajectory, CTC, CTC-GPR}
\end{axis}
\end{tikzpicture}
  \vspace{-0.4cm}
	\caption{Tracking performance for the first joint. The color of the CTC-GPR trajectory indicates the norm of the current feedback gains (red high, blue low). \label{fig:var_gains}}
\end{center}
\end{figure}
\begin{table}[b]
\vspace{-0.0cm}
\renewcommand{\arraystretch}{1.1}
\begin{tabularx}{\columnwidth}{@{}l *3{>{\centering\arraybackslash}X}@{}}
	\toprule
			&	CTC		&	\shortstack{Static\\CTC-GPR}	& \shortstack{Variable\\CTC-GPR}\\
			\midrule
	$\Verts{K_p}$	&	10		&	7.01	&	7.01 - 9.38 \\
    $\Verts{K_d}$	&	10   	&	6.06	&	6.06 - 9.38\\
        $\Verts{\qe\tran,\dqe\tran}_{L^2}$	& 	4.7281  &  1.8760   &	\textbf{1.5118}\\
    $\max(\Verts{\qe(t)})$	&	0.2420  &  0.1066   &	\textbf{0.0819}\\
    $\max(\Verts{\dqe(t)})$	&	0.2377  &  0.1234   &	\textbf{0.1002}\\
    \bottomrule
\end{tabularx}
\renewcommand{\arraystretch}{1}
  \caption{Comparision between CTC, CTC-GPR with static gains, and CTC-GPR with variable gains.\label{tab:comp}}
  \vspace{-0.1cm}
\end{table}
\subsection{Discussion}
\vspace{-0.1cm}
Both CTC-GPR approaches show a lower tracking error than the classic CT. The reason is that the CTC-GPR uses the mean function to compensate the unknown dynamics, such that the feedback gains can be lower in comparison to the CTC. Additionally, the variable CTC-GPR outperforms the static CTC-GPR for the position and velocity error because the gains are increased as soon as the trajectory leaves the training area. The result is that the tracking error is kept low and bounded even for areas where no training data is available. The additional benefits of low feedback gains for noise attenuation are shown in~\cref{fig:comp}. On the other side, the improved tracking performance of the CTC-GPR comes with the computationally demanding calculation of the predictive mean and marginal variance of the GP. The design of the variable gain functions and the effect on the closed loop performance is subject of future work.

\vspace{-0.3cm}
\section*{Conclusion}
\vspace{-0.2cm}
We propose a data-driven approach for high performance tracking control. It is based on a computed-torque control law where the feedback gains are adapted by the model fidelity of a data-driven model of the system. For this purpose, we use the mean prediction of the GPR to compensate the residual dynamics of the system and the variance to adapt the feedback gains. The main contribution is the determination of the tracking error of the closed loop system which is proven to be uniformly ultimately bounded and exponentially convergent to a ball with a given probability. The result shows the correlation between the bound of the tracking error, the uncertainty of the model and the feedback gains.
\vspace{-0.3cm}

\appendix 
\section{Proof of~\cref{lemma:boundederror}}
	\label{app:boundederror}
	The result is a consequence of~\cite[Theorem 6]{srinivas2012information} which concerns the one dimensional case. In this case, the training data is generated by a scalar function~$f\colon D		\to \R$ with~$f\in \mathcal{H}_k(D)$ on a compact set~$D\subset\R^n$. A GP is trained with~$m$ data points~$\mathcal D=\{\x^{\{i\}},y^{\{i\}}\}_{i=1}^m$ of 
	\begin{align}
		y&=f(\x)+\eta,\quad &y,\eta&\in\R,\x\in \R^n\label{for:gpgen}\\
		\eta&\sim\mathcal{N}(0,\sigma_1^2),&\sigma_1&\in\R_{>0}.
	\end{align}
	Then, recalling~\cite{srinivas2012information}, the model error~$\Delta\in\R$
	\begin{align}
		\Delta=\vert\mean(f\vert \x^*,\mathcal D)-f(\x^*)\vert
	\end{align}
	is bounded with a probability of at least~$\tilde\delta$ by
	\begin{align}
		\text{P}&\left\lbrace \forall \x^*\in D,\, \Delta\leq \vert \beta \Var^{\frac{1}{2}}(f\vert \x^*,\mathcal D)\vert\right\rbrace\geq \tilde\delta\label{for:upbound_scalar}
	\end{align}
	with~$\tilde\delta\in(0,1)$, where~$\beta\in\R$ is defined as
	\begin{align}
		\beta&=\sqrt{2\Verts{f}^2_k+300 \gamma \ln^3\left(\frac{m+1}{1-\tilde\delta}\right)}.
	\end{align}
	The variable~$\gamma\in\R$ is the maximum information gain 
	\begin{align}
		\gamma&=\hspace{-0.4cm}\max_{\bm{x}^{\{1\}},\ldots,\bm{x}^{\{m+1\}}\in D} I(y^{\{1\}},\ldots,y^{\{m+1\}} ; f)\\
		&= \hspace{-0.4cm}\max_{\bm{x}^{\{1\}},\ldots,\bm{x}^{\{m+1\}}\in D}\frac{1}{2}\log \vert I+\sigma_1^{-2}K_{\Phi_1}(\x,\x^\prime)\vert
	\end{align}
	with covariance matrix~$K_{\Phi_1}(\x,\x^\prime)$ and the input elements~$\x,\x^\prime\in\{\bm{x}^{\{1\}},\ldots,\bm{x}^{\{m+1\}}\}$. In the multidimensional case of~\cref{lemma:boundederror}, we use a GP 		for each dimension of~$\tilde{\bm \tau}(\bm p)$ as shown in~\cref{for:multigp}. For the calculation of~\cref{for:upbound}, assume the two sets
	\begin{align}
		\Pi_A&\hspace{-0.2em}=\hspace{-0.2em}\left\lbrace \forall\bm p\in D,\vert\mean(\tilde{\tau}_j\vert \bm{p},\mathcal D)-\tilde{\tau}_j(\bm p)\vert\leq \beta_j\var^{\frac{1}{2}}(\tilde{\tau}_j\vert \bm{p},\mathcal D)\right\rbrace\notag\\
		\Pi_B&\hspace{-0.2em}=\hspace{-0.2em}\left\lbrace\forall\bm p\in D,\Verts{\Mean(\tilde{\bm \tau}\vert \bm{p},\mathcal D)-\tilde{\bm \tau}(\bm p)}\hspace{-0.2em}\leq\hspace{-0.2em} \Verts{\bm \beta\tran \Var^{\frac{1}{2}}(\tilde{\bm \tau}\vert \bm{p},\mathcal D)}\right\rbrace
	\end{align}
	with the multidimensional extension~$\bm\beta,\bm\gamma\in\R^n$
	\begin{align}
		\beta_j&=\sqrt{2\Verts{\tilde{\tau}_j}^2_k+300 \gamma_j \ln^3\left(\frac{m+1}{1-\delta^{1/n}}\right)}\\
		\gamma_j&=\max_{\bm{p}^{\{1\}},\ldots,\bm{p}^{\{m+1\}}\in D}\frac{1}{2}\log \vert I+\sigma_j^{-2}K_{\Phi_j}(\x,\x^\prime)\vert\notag\\
		\x,\x^\prime&\in\left\lbrace\bm{p}^{\{1\}},\ldots,\bm{p}^{\{m+1\}}\right\rbrace,j\in\{1,\ldots,n\}.\notag
	\end{align}
	Due to the fact that~$\tilde{\bm \tau}$ is assumed to be uncorrelated~\cref{for:multigp}, the conditional probability for the set~$\Pi_A$ is lower bounded by $\text{P}\left\lbrace \Pi_A \right\rbrace\geq \delta=\tilde\delta^n$. With the monotony property of the probability measure~$P$ and since~$\Pi_A\subseteq\Pi_B$ holds,~\cref{for:upbound} provides an upper bound for the norm of the model error with a probability of at least~$\delta\in(0,1)$.\qed
	\vspace{-0.2cm} 
\section{Negative definiteness of $M$ in~\cref{for:dotV}}
\label{app:mneg}
	According to Schur's lemma, the matrix~$M$ of~\cref{for:dotV} is negative definite if $M_{11}=-K_d(\Var_d)+\varepsilon \hat H$ and
	\begin{align}
		S&=-\varepsilon K_p(\Var_p)+\frac{\varepsilon^2}{4}(K_d(\Var_d)-\hat C\tran)\notag\\
		&\phantom{=}(K_d(\Var_d)-\varepsilon \hat H)^{-1}(K_d\tran(\Var_d)- \hat C)
	\end{align}
	are negative definite, where~$M_{11}\in\R^{n\times n}$ is the upper left block of~$M$ and~$S\in\R^{n\times n}$ is the Schur complement. Since~$K_d,\hat H$, and~$K_p$ are positive definite and bounded,~$\varepsilon$ can be chosen sufficiently small to obtain the negative definiteness of~$M_{11}$. The second summand of the Schur complement~$S$ is quadratic in~$\varepsilon$ and positive definite, while the first summand is linear in~$\varepsilon$ and negative. Thus, for every~$\q,\dq\in\R^n$, an~$\varepsilon$ can be found which guarantees the negative definiteness of the Schur complement. Therefore, there exists an~$\epsilon>0$, so that matrix~$M$ is negative definite. \qed
\bibliographystyle{plain}        
\bibliography{mybib}           

\end{document}